# Nonparametric Spatio-Temporal Joint Probabilistic Data Association Coupled Filter and Interfering Extended Target Tracking

**Behzad Akbari[1], Member, IEEE, Haibin Zhu[2], Ya-Jun Pan[1], Senior Members, IEEE, and R. Tharmarasa[3]**

[1] Department of Mechanical Engineering, Dalhousie University, Halifax, Nova Scotia, Canada, B3H 4R2

[2] Collaborative Systems Laboratory (CoSys Lab), Department of Computer Science and Mathematics, Nipissing University, North Bay, Ontario, Canada, P1B 8L7

[3.] Electrical and Computer Engineering Department, McMaster University, Hamilton, Ontario, Canada, L8S 4K1

Corresponding author: Behzad Akbari (e-mail: bh888652@dal.ca).

This work was partly supported by the Natural Sciences and Engineering Research Council, Canada (NSERC) under Grant RGPIN-2018-04818 and RGPIN-2021-02406.

**ABSTRACT** Extended target tracking estimates the centroid and shape of the target in space and time. In various situations where extended target tracking is applicable, the presence of multiple targets can lead to interference, particularly when they maneuver behind one another in a sensor like a camera. Nonetheless, when dealing with multiple extended targets, there's a tendency for them to share similar shapes within a group, which can enhance their detectability. For instance, the coordinated movement of a cluster of aerial vehicles might cause radar misdetections during their convergence or divergence. Similarly, in the context of a self-driving car, lane markings might split or converge, resulting in inaccurate lane tracking detections. A well-known joint probabilistic data association coupled (JPDAC) filter can address this problem in only a single-point target tracking. A variation of JPDACF was developed by introducing a nonparametric Spatio-Temporal Joint Probabilistic Data Association Coupled Filter (ST-JPDACF) to address the problem for extended targets. Using different kernel functions, we manage the dependency of measurements in space (inside a frame) and time (between frames). Kernel functions are able to be learned using a limited number of training data. This extension can be used for tracking the shape and dynamics of nonparametric dependent extended targets in clutter when targets share measurements. The proposed algorithm was compared with other well-known supervised methods in the interfering case and achieved promising results.

## I. INTRODUCTION

One of the main assumptions in common target tracking problems is that each target can generate at most one measurement per scan. These types of problems are called single-point tracking problems. With the recent developments in high-resolution sensors, tracking the shape of the target in addition to its kinematics is attracting attention in ground, water, and air surveillance applications. The process of tracking the shape and kinematics of an object simultaneously is called Extended Target Tracking (ETT) [1]. ETT has become an essential part of many autonomous and self-driving systems. The main focus is on the improvement of precision, robustness, and safety. The focus of ETT has been mostly on measurement models, shape estimation, and data association. A typical approach to an ETT problem is to estimate the kinematics of the center of the object (CoO) and model the extent as a function that is sometimes unknown and nonlinear, using a sequence of noisy measurements. For ETT, various types of sensors, such as cameras, radars, and LiDAR (light detection and ranging), are employed to fuse information for detecting and tracking objects and obstacles covered with different types of clutter. In some early published papers, because of difficulties in estimating the complete shape, the relaxed version employed a rectangular, ellipsoidal, or cubic frame to illustrate the object's centroid, size, or direction. Some of the early works are summarized in detail in [2]. Most of these techniques are underpinned by random finite sets (RFS). For instance, PHD filter [3], and its derivatives operate on this principle. Random Matrices (RM) was the first Bayesian method for tracking extended targets [4]. The formulation of the RM technique assumes that the extended target is ellipsoidal, but that is not always true. To describe and estimate the non-ellipsoidal shape of an extended target, several techniques, such as the multi-ellipsoid RM model [5] and random hyper-surface model [6]- [7], have been proposed. Most of the previous methods were parametric, not trainable, and relaxed the shape of the target with elliptical or rectangular models. In [8], a Gaussian Process (GP) was used for the first time to model the extent of the target as a continuous function. A GP is a distribution over an unknown and nonlinear function in the continuous domain. The observed values of these functions (measurements) can be used to predict the values at unobserved points. Since the GP is a batch process and cannot be used in real-time applications, [8] proposes a recursive version of GP, where the change in the shape is modeled using a forgetting factor model. The idea of evolving the shape with a kernel-based GP platform developed more in [9] by introducing the recursive Spatio-Temporal Gaussian Process (STGP), in which it considered separable covariance functions for spatial and temporal dependency. In [9], a factorization of the power spectral density function of the covariance function is employed to track the extent evolving in time. Additionioinally, the idea of converting a covariance function to a state-space model solvable with a smoother similar to [10] were used. The GP is nonparametric, compatible with linear Markov models, and can be used ideally for star convex single extended targets without considering outliers and interfering.

In the case where multiple extended targets are in the scene, associating the proper particle measurements to the target in clutter "data association" is a challenge. A Joint Probabilistic Data Association Filter (JPDAF) can handle multiple single-point targets that may share measurements. In the original JPDAF, we assume that the target states and, thus, the target-originated measurements are independent. For the dependent single-point targets, a statistical dependence of their estimations can be taken into account by calculating the state cross-covariances. The resulting algorithm is called the Joint Probabilistic Data Association Coupled Filter (JPDACF) [11]. A motion model in the traditional JPDACF is parametric and not trainable and must mostly be tuned precisely at the design stage.

How can we facilitate the training of JPDACF? How can we carry out measurements for the association of extended targets? Moreover, how can we harness the benefits of interdependence among a cluster of extended targets to enhance association accuracy? To tackle these challenges, this paper introduces a kernel-infused JPDACF and employs a novel measurement model that is rooted in joint probabilistic data association. This model combines filtering with association to effectively track and associate extended targets, addressing the issues mentioned above. The parameters of the filter are learned from the limited number of training data. For modeling dependency between targets, similar to [16] and [12] we used a coregionalization matrix showing kernels' similarity. The algorithm generally enables us to track the shape of multiple dependent extended targets in a cluttered situation where targets share measurements.

The following are the main contributions of our work:

1. We created an ST-JPDACF method to track extended dependent targets in both space and time;
2. We developed a nonparametric model based on a JPDACF for extended target data association in space, allowing for the prediction of measurements and measurement likelihood in a cluttered environment;
3. We developed a training algorithm for space and time kernels to determine transition matrices for stacked states and covariance;
4. We applied the ST-JPDACF to track multiple dependent lane markings in a cluttered environment.

The remaining content of the paper is structured as follows: In the beginning, we elucidate our problem formulation in section II. This includes an explanation of how a kernel-based semi-supervised approach can be employed to train the JPDACF, determine measurement likelihood for data association, and infer extended targets. In the subsequent section, section III, we present several applications of the suggested technique in the context of lane line tracking.

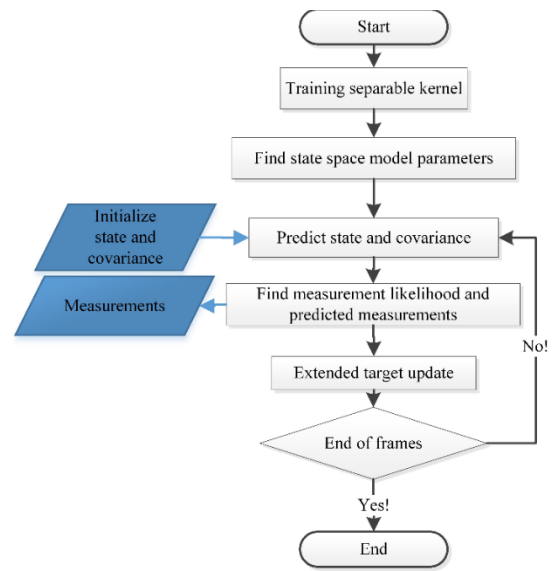

**FIGURE 1. General pipeline for tracking ET using ST-JPDACF.**

**FIGURE 2. TABLE I. Abbreviations adopted in this manuscript**

| Abbreviation | Description |
|---|---|
| ETT | Extended Target Tracking [1] |
| RFS | Random Finite Sets [3] |
| PHD | Probability Hypothesis Density Filter [3] |
| RM | Random Matrices [5] |
| GP | Gaussian Process [12] |
| JPDA | Joint Probabilistic Data Association [11] |
| JPDACF | Joint Probabilistic Data Association Coupled Filter [11] |
| ST-JPDACF | Spatio Temporal JPDACF |
| RTS | Rauch-Tung-Strieble [13] |
| STGP | Spatio Temporal Gaussian Process [10] |
| SCNN | Spatial Convolutional NN [14] |
| RBF | Radial Basis Function |
| BFGS | Broyden-Fletcher-Goldfarb-Shanno [15] |
| SDE | Stochastic Differential Equation |
| IPM | Inverse Perspective Mapping |

## II. PROBLEM FORMULATION

In this section, an ST-JPDACF is introduced as a semisupervised method to model the evolution of a continuous stochastic function in both space and time. The general pipeline of ST-JPDACF is illustrated in Fig.1. The idea of nonparametric ST-JPDACF comes from the application of kernel models in pattern recognition and machine learning. In kernel-based models, the dependency between samples is summarized in a function called the kernel. The kernel is a symmetrical and not negative function with enough information to explain the dependency of samples in a stochastic function. This model is semisupervised because the kernel hyperparameters can be learned from the training data. This paper uses a nonparametric ST-JPDACF with separable already-trained space and time kernels. Trained kernels will be used to find the measurement likelihood and transition matrices. ST-JPDACF can model a kernel-based stochastic function for a system evolving in both space and time, specified by:

$$f(u,t) \sim \mathcal{ST}\_JPDACF\big(\mu(u,t), \kappa(u,u';t,t')\big), \tag{1}$$

$$z_i = \mathcal{H}_i f(u_i, t_i) + v_i,$$

where $u_i$ is an index point for the state value function $f_i(u_i, t)$ , $t$ is the time, and $\mu(u,t)$ and $\kappa(u,u';t,t')$ represent the mean and the covariance function of the stochastic function. Eq. (2) represents any stochastic function as a differential equation. To use the Markov property for solving this differential equation, we can consider a sufficient number of time derivatives and state space model as Eq.(3) [10].

$$a_n \frac{d^n f(u,t)}{dt^n} + \dots + a_1 \frac{df(u,t)}{dt} + a_0 f(u,t) = \mathbf{w}(u,t), \tag{2}$$

$$d\mathbf{f} = \mathcal{A}\mathbf{f} + \mathbf{L}\mathbf{w}(u,t), \tag{3}$$

where $\mathbf{f} = [f, df/dt, \dots d^{s-1}f/dt^{s-1}]^\top$ , and $\mathbf{w}(u,t)$ shows the smoothness of the shape, which can be a Wiener process. Bayesian Kalman Filtering is a linear way to solve this equation when clutter is not an issue, and targets are not interfering. However, filtering provides the forward-time posteriors, and to acquire the full posterior, a linear smoother also needs to be involved. First, [10] has shown that by considering some assumptions, the transition matrices can be computed by an inverse Fourier transform of the power spectral density of the temporal kernel. The covariance kernel was considered separable to simplify the result because the correlation structure obeys different dependencies through space and time.

$$\kappa(u,u';t,t') = \kappa_u(u,u')\kappa_t(t,t'), \tag{4}$$

where $\kappa_u(u,u')$ and $\kappa_t(t,t')$ are the spatial and temporal covariance functions. Considering the conditions in [10], the temporal stochastic process Eq. (3) can be equivalently

represented by an infinite dimensional dynamical system given below:

$$\frac{\partial \mathbf{f}(u,t)}{\partial t} = \mathcal{A}_t \mathbf{f}(u,t) + \mathbf{L}_t \mathbf{w}_t(u,t), \tag{5}$$

$$\mathcal{A}_t = \begin{bmatrix} 0 & 1 & & \\ & \ddots & \ddots & \\ & & 0 & 1 \\ -a_0 & \dots & -a_{s-2} & a_{s-2} \end{bmatrix}, \quad \mathbf{L}_t = \begin{bmatrix} 0 \\ \vdots \\ 0 \\ 1 \end{bmatrix}, \tag{6}$$

where $\mathbf{f}(u,t) = [f, \partial f/\partial t, \dots \partial f^{s-1}/\partial t^{s-1}]^\top$ is a state at time $t$, consisting of a function and a suitable number of its time derivatives, $\mathcal{A}_t$ is the continuous state transition with a matrix form, $\mathbf{L}_t$ represents the noise effect and $\mathbf{w}_t(u,t)$ is a zero mean continuous-time white process noise. In a finite collection of index points $\mathbf{u} = \{u_i\}_{i=1}^N$ and discrete-time $t_k$ the system model can be defined as Eqs. (7) and (8):

$$\mathbf{f}(\mathbf{u}, t_k) = \mathbf{F}_k \mathbf{f}(\mathbf{u}, t_{k-1}) + \mathbf{L}_k \mathbf{w}_k(\mathbf{u}), \quad \mathbf{w}_k \sim \mathcal{N}(0, \mathbf{Q}_k(\mathbf{u}, \mathbf{u}'; T_s)), \tag{7}$$

$$\mathbf{z}_k = \mathbf{H}_k \mathbf{f}(\mathbf{u}, t_k) + \mathbf{v}_k, \quad \mathbf{v}_k \sim \mathcal{N}(0, R). \tag{8}$$

The corresponding states for an extended target would be :

$$\mathbf{f}(\mathbf{u}, t_k) = [[f(u_1, t_k), \dot{f}(u_1, t_k), \dots], [f(u_2, t_k), \dot{f}(u_2, t_k), \dots], \tag{9}$$
$$\dots, [f(u_{Nf}, t_k), \dot{f}(u_{Nf}, t_k), \dots]]^\top,$$

Using the Kronecker product, the discrete transition matrices are computed as: $\mathbf{F}_k = I \otimes e^{\mathcal{A}T_s}$, $T_s = t_k - t_{k-1}$ . The separable process noise covariance matrix for $N$ index points has a spatial structure as follows:

$$\mathbf{Q}(\mathbf{u}, \mathbf{u}'; T_s) = K_u(\mathbf{u}, \mathbf{u}')[I_N \otimes \mathbf{Q}_k], \tag{10}$$

$$\mathbf{Q}_k = \mathbf{P}_\infty - \mathbf{F}_k \mathbf{P}_\infty \mathbf{F}_k^\top.$$

The initial covariance is $\mathbf{P}_0 = \mathbf{P}_\infty$ . Associating particle measurements to the extended targets is very challenging when multiple targets interfere in a cluttered environment. The nonparametric JPDACF [11] was utilized in this paper, as opposed to GP regression, which differs from the methods used in references [9] and [16]. In fact, JPDACF will find the extent of the objects that are considered dependent functions $\mathbf{f}(\mathbf{u}, t_k)$. To deal with clutter, rather than choosing the most likely assignment of measurements to a target (or declaring the target not detected or measurement to be a false alarm), the JPDACF calculates a real-time probability of each validated measurement to be attributable to the target of interest. This probabilistic information is used in a tracking algorithm to deal with the measurement origin uncertainty in a cluttered environment. The parameters of a JPDACF and transition matrixes could be trained based on training data. The result of the nonparametric JPDACF algorithm in the space domain would be pseudo measurements and likelihood. Section II-A comprehensively explains the process for determining the measurement likelihood. With pseudo measurements and likelihoods available, this model facilitates sequential inference using a multivariate recursive Bayesian Filter. The computation order of this process is $O(MN^3T)$, when M is the number of particle measurements, $N$ represents the state dimension, and $T$ represents a time step.

## A. Measurements likelihood

In our application, JPDACF will augment, summarise and unify the available real-time measurements in the form of pseudo measurements. To identify the kernel hyperparameters and corresponding state space transition matrices for a specific extended target, a limited amount of training data in the spatial domain is utilized. The extended targets are viewed as a set of stochastic functions that are represented by trajectories. In the case of a high dimensional object, its form can be characterized by a closed loop corresponding to the contour of its surface. However, within this research, we exclusively addressed straightforward curved entities and did not address periodic functions involving closed loops. At each step, the nonparametric JPDACF is utilized to process the received measurements and associate them to create pseudo measurement. In this paper, the spatial kernel for two individual indexes $(u, u')$ is considered a square exponential or Radial Basis Function (RBF) kernel ( Eq. 11). The distinction between the RBF and Matern kernels is apparent in Fig.2.

$$\kappa(u, u')_{RBF} = \sigma^2 \exp(-\frac{1}{2}\frac{(u-u')^2}{\ell^2}).\qquad(11)$$

Hyperparameters of the kernel $\sigma$, $\ell$, and the dependency matrix $B_D$ are learned from the training data, Algorithm 2. The parameters of the nonparametric JPDACF come from the power spectral density of the trained spatial kernel function.

$$S(\omega)_{RBF} = \sigma^2\sqrt{2\pi}/\exp(-\frac{\ell^2(\omega)^2}{2}).\qquad(12)$$

The RBF spectral density is not a rational function, but we can approximate it with such a function. By denoting $\zeta = \frac{1}{2l^2}$ and $N$ the approximation order using the Taylor series of the exponential function :

$$S(\omega)_{RBF} \approx \sigma^2 N! (4\zeta)^N \sqrt{\frac{\pi}{\zeta}} (\frac{1}{\sum_{n=0}^{N}\frac{N!(4\zeta)^{N-n}}{n!}\omega^{2n}}).\qquad(13)$$

In order to find the transfer function, we factorize the denominator $\sum_{n=0}^{N}\frac{N!(4\zeta)^{N-n}}{n!}\omega^{2n}$ into stable and unstable parts. Because $N$ is even, the coefficient of $\omega^{2n}$ is 1, and we can form a polynomial function and find negative real part roots for $\mathcal{A}_0, \mathcal{A}_1, \ldots \mathcal{A}_N$ and make the $N$-dimensional state-space model as:

$$\mathcal{A}_t = \begin{bmatrix} 0 & 1 \\ -\mathcal{A}_0 & -\mathcal{A}_1 \end{bmatrix}, \quad L_t = \begin{bmatrix} 0 \\ 1 \end{bmatrix}.\qquad(14)$$

The spectral density of the Gaussian white noise process $w(t)$ is $\mathbf{Q}_c = \sigma^2 N! (4\zeta)^N \sqrt{\frac{\pi}{\zeta}}$ and the measurement model matrix is $\mathbf{H}_t = [1\ 0]$. The stationary state corresponds to the state that the model stabilizes to infinity ($f_\infty \sim \mathcal{N}(0, P_\infty)$). The stationary state corresponds to the state that the model stabilizes to infinity. It can be represented by the stationary covariance of $f(t)$ that is the solution to:

$$\frac{d\mathbf{P}_\infty}{dt} = \mathbf{F}\mathbf{P}_\infty + \mathbf{P}_\infty\mathbf{F}^\top + \mathbf{L}q_t\mathbf{L}^\top = 0.\qquad(15)$$

The stationary state is invariant to the input location choice and describes the state the process defaults to. Therefore, for stationary models, the initial (prior) state is given by the stationary state covariance, $\mathbf{P}_0 = \mathbf{P}_\infty$. For $D$ unknown number of tracks $\Gamma_t$ , the recursive algorithm for measurement likelihood uses the following steps:

*1. Track Initialization:* Define a track $\Gamma_t$ for the target $t$, a tuple of state, covariance, and associated measurements $\Gamma_t = \{x, P, \mathbf{z}_t\}$ . For all values of the index vector $u$, measurements are taken. If there are some tracks, go to Step 2; otherwise, initialize a new track based on the new measurements.

*2. Coupling:* We need to stack states to consider dependency between trajectories. For $D$ number of tracks in step $k$ , stacked states, covariance, and filter matrices are as follows:

$$\bar{x}_\varkappa = [x_\varkappa^1, x_\varkappa^2, \ldots, x_\varkappa^D]^\top, \quad \bar{P}_\varkappa = B_D \otimes P_\varkappa,\qquad(16)$$

$$\bar{F}_\varkappa = I_D \otimes F_\varkappa, \quad \bar{Q}_k = B_D \otimes Q_\varkappa \quad \bar{R}_k = I_D \otimes R_\varkappa.$$

Notice that $\varkappa$ is step time and $B_D$ is a coregionalization matrix [17]. $B_D$ is a square matrix of size equal to the number of targets, where the off-diagonal elements represent the extent of dependency between targets for each corresponding point. In the independent situation, $B_D = I_D$ . Finding hyperparameters from the training data $\{\mathbf{u}_d, \mathbf{z}_d\}_{d=1}^D$ is similar to a regular GP, we only need to add extra parameters for the coregionalization matrix $B$. The approach normally taken minimizes the log-likelihood of marginal likelihood via optimization techniques such as maximizing marginal likelihoods or gradient descent methods like Broyden–Fletcher–Goldfarb–Shanno(BFGS) algorithm [18], [19]. Since $D$ is unknown, the integrated version of the nonparametric JPDACF is employed to include a target management system for unknown targets [20]( Algorithm 1).

*3. Prediction:* We use the prior information coming from the spatial kernel function for prediction. We imagine that the spatial kernel is already known from the training data. The inverse Fourier transform of the power spectral density of the kernel gives transition matrixes similar to [10]. Considering the last estimates ($\hat{\bar{x}}_{\varkappa-1|\varkappa-1}, \bar{P}_{\varkappa-1|\varkappa-1}$), the new states and covariance can be found to be similar to a normal Bayesian filter prediction, and the results are ($\hat{\bar{x}}_{\varkappa|\varkappa-1}, \bar{P}_{\varkappa|\varkappa-1}$) (Eq. 17). Note that the predicted covariance is the cross-covariance with off-diagonal blocks.

$$\hat{\bar{x}}_{\varkappa|\varkappa-1} = \bar{F}_{\varkappa-1}\bar{x}_{\varkappa-1|\varkappa-1},\qquad(17)$$

$$\bar{P}_{\varkappa|\varkappa-1} = \bar{F}_{\varkappa-1}\bar{P}_{\varkappa-1|\varkappa-1}\bar{F}_{\varkappa-1}^\top + \bar{Q}_{\varkappa-1}.$$

*4. Gating:* Even though we use the stacked inputs, a measurement validation must be performed separately for each target. To do so, we need to find $S_\varkappa^d$ as the $n_z \times n_z$ sub-block diagonal of $\bar{S}_\varkappa$ for a single predicted measurement $\hat{z}_\varkappa^d$. Based on the normal gating and Mahalanobis distance, the measurement $z_\varkappa^j$ can be validated if and only if:

$$[z_\varkappa^j - \hat{z}_\varkappa^d][S_\varkappa^d]^{-1}[z_\varkappa^j - \hat{z}_\varkappa^d]^\top < \gamma,\qquad(18)$$

where $\gamma$ is an appropriate threshold.

5. *Computing the association events probabilities*: The conditional probability for a joint association event needs to involve a clutter model and target interference. Similar to a regular JPDACF, a possibility of associating an event $A_\varkappa$ in the time step $k$ for measurements $Z^\varkappa = \{z_j\}_{j=1}^{\varkappa}$ comes from Bayes formula in Eq.19:

$$P\{A_\varkappa | Z^\varkappa\} = P\{A_\varkappa | z_\varkappa, m_\varkappa, Z^{\varkappa-1}\}$$
$$= \frac{1}{c} p[z_\varkappa | A_\varkappa, m_\varkappa, Z^{\varkappa-1}] P\{A_\varkappa | Z^{\varkappa-1}, m_\varkappa\}, \quad (19)$$

$$A_\varkappa = \cap_{j=1}^{m} A_{j t_j}(\varkappa).$$

Note that $A_{j t_j}(\varkappa)$ is the event that measurement $j$ at time $\varkappa$ originated from target $t$, $j = 1, .., m_\varkappa$, $t = 0,1, \ldots, N_T$; $t_j$ is the index of the target to which measurement $j$ is associated in the event under consideration, and $N_T$ is the known number of targets. Having $m_\varkappa$ events, the density of the event $A_\varkappa$ given validated measurements $Z^{\varkappa-1}$ is written as:

$$P\{A_\varkappa | Z^{\varkappa-1}, m_\varkappa\} = P\{A_\varkappa | \delta_A, \phi_A, m_\varkappa\} P\{\delta_A, \phi_A | m_\varkappa\}, \quad (20)$$

where $\phi_A$ is the number of false measurements, and the binary variable $\delta_t$ is the detection indicator (equal to one if a target $t$ is assumed to be detected in the event $A_\varkappa$). The last expression above can be rewritten as:

$$P\{A_\varkappa | Z^{\varkappa-1}, m_\varkappa\} = \frac{\mu_{F,\phi}}{P\{m_\varkappa\}} (\rho_{m_\varkappa - \phi_A}^{m_\varkappa})^{-1} \prod_t (P_D^t)^{\delta_t} (1 - P_D^t)^{1-\delta_t} \quad (21)$$

$$= \frac{\phi!}{c \, m_\varkappa!} \mu_{F,\phi} \prod_t (P_D^t)^{\delta_t} (1 - P_D^t)^{1-\delta_t},$$

where $(\rho_{m_\varkappa - \phi_A}^{m_k})$ is an arrangement of $m_\varkappa$ over $(m_\varkappa - \phi_A)$ measurements, $\mu_{F,\phi_A}$ is the prior probability mass function (Poisson pmf) of the clutter model, $P_D$ is the probability of detection, and $c$ is the normalization constant. The states of the target are dependent on each other, and the conditional measurement likelihood, in this case, cannot be reduced to the marginal form:

$$p[z_\varkappa | A_\varkappa, m_\varkappa, Z^{\varkappa-1}] = V^{-\phi_A} f_{t_{j1}, t_{j2}..}[z_{j,\varkappa}, j : \tau_j = 1], \quad (22)$$

where $V$ is the volume of the surveillance region in which the measurements not associated with a target are to be assumed uniformly distributed and $f_{t_{j1}, t_{j2}..}$ are the joint probability density function of the measurements of the targets under consideration, and $t_{j1}$ is the target to which $z_{j1,\varkappa}$ is associated with Event $A$. $\tau_j$ is the target association indicator for measurement $j$ in the event $A_\varkappa$.

6. *Update*: In a coupled filter, the joint probabilities are not reduced to the marginal form. Instead, these joint probabilities are used directly in a coupled filter as follows:

$$\hat{x}_{\varkappa | \varkappa} = \hat{\hat{x}}_{\varkappa | \varkappa-1} + \bar{G}_\varkappa \bar{v}_\varkappa, \quad (23)$$

where

$$\bar{v}_\varkappa = \sum_A P\{A_\varkappa | Z^\varkappa\} \bar{v}_{A,\varkappa}, \text{ and } \bar{v}_{A,\varkappa} = [\bar{z}_{\varkappa,A} - \hat{z}_{\varkappa | \varkappa}], \quad (24)$$

$$\bar{z}_{\varkappa,A} = [z_{j1,A}^\varkappa, z_{j2,A}^\varkappa, \ldots, z_{jD,A}^\varkappa]^\top, \quad (25)$$

and the subscript $_{jt,A}$ is the index for the measurement associated with target $t$ in Event $A$ at time step $\varkappa$. The conditional measurement matrix $H_A(\varkappa)$ needs to take care of the cases where some of the targets are not detected.

$$\bar{H}_A(\varkappa) = Diag[\delta_A^1, \delta_A^2, \ldots, \delta_A^D] \otimes H(\varkappa), \quad (26)$$

where $[\delta_A^1, \delta_A^2, \ldots, \delta_A^D]^\top$ is a target detection indicator vector. The filter gain in (19) is computed by inverting the covariance of residual $\bar{S}_\varkappa = \bar{H}_\varkappa \bar{P}_{\varkappa | \varkappa-1} \bar{H}_\varkappa^\top + \bar{R}_\varkappa$:

$$\bar{G}_\varkappa = \bar{P}_{\varkappa | \varkappa-1} \bar{H}_\varkappa^\top [\bar{H}_\varkappa \bar{P}_{\varkappa | \varkappa-1} \bar{H}_\varkappa^\top + \bar{R}_\varkappa]^{-1}. \quad (27)$$

The predicted stacked measurement vector is:

$$\hat{z}_{\varkappa | \varkappa-1} = \bar{H}_\varkappa \hat{x}_{\varkappa | \varkappa-1}. \quad (28)$$

Having all the matrices above, the state covariance update is as follows:

$$\bar{P}_{\varkappa | \varkappa} = \beta_{0,\varkappa} \bar{P}_{\varkappa | \varkappa-1} + [1 - \beta_{0,\varkappa+1}] \bar{P}_{\varkappa | \varkappa}^C + \tilde{P}_{\varkappa-1}. \quad (29)$$

With the probability $\beta_{0,\varkappa} \triangleq P\{A_0 | Z^\varkappa\}$, none of the measurements is correct, and there is no update on the state estimate. With a probability $1 - \beta_{0,\varkappa}$, the correct measurement is available, and the updated covariance is $\bar{P}_{k|k}^C$ as:

$$\bar{P}_{\varkappa | \varkappa}^C = \bar{P}_{\varkappa | \varkappa-1} - \bar{G}_{\varkappa-1} \bar{S}_{\varkappa-1} \bar{G}_{\varkappa-1}^\top. \quad (30)$$

However, since it is not known which of the $m_k$ validated measurements are correct, the term $\tilde{P}_{\varkappa-1}$, which is positive semidefinite, increases the covariance of the updated state. The spread of the innovations term is:

$$\tilde{P}_{\varkappa-1} \triangleq \bar{G}_\varkappa \left[ \sum_A P\{A_\varkappa | Z^\varkappa\} \bar{v}_{A,\varkappa} \bar{v}_{A,\varkappa}^\top - \bar{v}_\varkappa \bar{v}_\varkappa^\top \right] \bar{G}_\varkappa^\top. \quad (31)$$

7. *Making new tracks*: After updating tracks, we may have some non-associated measurements that need to be considered for new tracks. See Algorithm 1.

---

**Algorithm 1. Integrated Nanparametric JPDACF**

**Input:** $\mathbf{Z}, k(u, u'), B$
**Output:** $\Gamma = \{(\Gamma_t, x, \Gamma_t.P, \Gamma_t.\mathbf{z}_t)\}_{t=0}^D$ #set of tracks
1: $\Gamma = \{\}$  #Initialize tracks to Null
2: **for** $\varkappa$ **in** range(u)  **do**
3:    $\mathbf{z}_\varkappa \leftarrow$ getMeas($\mathbf{Z}, \varkappa$)
4:    **if** $\Gamma.size > 0$ **then**
5:      $\bar{x}, \bar{F}, \bar{H}, \bar{P}, \bar{Q}_c \leftarrow$ Coupling $(\Gamma_\varkappa, \kappa(.), B)$
6:      $\hat{\bar{x}}, \hat{\bar{P}}, \hat{\bar{z}} \leftarrow$ Prediction $(\bar{F}, \bar{H}, \bar{x}, \bar{P}, \bar{Q}_c)$
7:      $\bar{V}\mathbf{z} \leftarrow$ gating $(\hat{\bar{x}}, \hat{\bar{P}}, \hat{\bar{z}}, \mathbf{z}_\varkappa)$
8:      $NonAssMeas \leftarrow \{\{\mathbf{z}\} - \{\bar{V}\mathbf{z}\}\}$
9:      $\Gamma_\varkappa \leftarrow$ JPDACF_Update $(\hat{\bar{x}}, \hat{\bar{P}}, \bar{V}\mathbf{z})$
10:      $\Gamma \leftarrow$ AddNewTracks $(\{\Gamma\}, NonAssMeas)$
11:    **else**
12:      $\Gamma \leftarrow$ AddNewTracks $(\{\Gamma\}, \mathbf{z})$
13:    **endif**
14: **endfor**
15: **return** $\Gamma$

---

### B. Fixed Lag Coupled RTS Smoother

In the state estimation, "smoothing" is a process where current measurements are used to improve estimates of past states. In fact, smoothing estimates the backward posterior density. The general case of smoothing used in real-time applications is called a fixed lag smoother. A fixed lag smoother smoothes the trajectory for an interval of $s$ states. Fixed interval smoothing requires a backward iteration after completing the (forward) filtering. When the targets are coupled, the results of filtering are in the stacked form, $\overline{x}_{k|k}, \overline{x}_{k+1|N}, \overline{P}_{k|k}, \overline{P}_{k+1|k}$ ($k = k, k-1, \ldots, k-s$), and need to be stored in the backward recursion for future use. In fixed interval smoothing, when the smoothing depth ($s$) is fixed, the smoother gains $\overline{G}_k$, the smoothed state $\widetilde{\overline{x}}_k$, and the state error covariance $\widetilde{\overline{P}}_k$ are recursively estimated using the Rauch-Tung-Striebel (RTS) recursion [13, 21] as:

$$\overline{G}_k = \overline{P}_{k|k}\overline{F}^\top(\overline{P}_{k+1|k})^{-1}, \tag{32}$$
$$\widetilde{\overline{x}}_{k|N} = \overline{x}_{k|k} + \overline{G}_k[\widetilde{\overline{x}}_{k+1|N} - \overline{x}_{k+1|k}],$$
$$\widetilde{\overline{P}}_{k|N} = \overline{P}_{k|k} + \overline{G}_k[\widetilde{\overline{P}}_{k+1|N} - \overline{P}_{k+1|k}]\overline{G}_k^\top.$$

The smoother is initialized at the current time step $k$ as $\widetilde{\overline{x}}_{k|N} = \overline{x}_{k|k}$ and $\widetilde{\overline{P}}_{k|N} = \overline{P}_{k|k}$. The result of smoothing function $f(.)$ at time $k$ is summarized in the mean and covariance $\widetilde{\overline{x}}_{k|N}$ and $\widetilde{\overline{P}}_{k|N}$, conditioned to the measurements $z_1, z_2, \ldots z_N$.

### C. Extended Target Inference

The result of the nonparametric JPDACF algorithm, section II-A, has several tracks, including stats, covariances and associated measurements. We consider the capital notations as a vector value function representing the state of an extended target, e.g., $f(u^f)$ is a vector value function for index vector $u^f$. To implement independent or dependent targets in a similar framework, we stack all the latent functions and consider them a single state vector. The notation of latent function $f(u^f)$ is changed to $x$ to be more similar to the state space models.

The dynamics of the extent are designed as separable kernels, which satisfy the conditions described in Eq. 4. The evolution of stacked extended target states is modeled as follows:

$$\overline{x}_{k+1} = \overline{F}_k\overline{x}_k + \overline{w}_k, \tag{33}$$
$$\overline{w}_k \sim \mathcal{N}(0, \overline{Q}_k),$$
$$\overline{x}_k = [f_1(u^f), f_2(u^f), \ldots, f_D(u^f)]^\top, \tag{34}$$

where $\overline{x}_k$ is the stacked states for coupled extended targets, each of which corresponds to $N$ distinct index points $u^f = \{u_1^f, u_2^f, \ldots u_N^f\}$. $\overline{F}_k$ is a stacked version of the state transition matrix and $\overline{w}_k$ is a zero-mean white Gaussian noise with covariance $\overline{Q}_k$ for the stacked states. The cardinality and value of the index points $u^f$ are related to the precision and shape of the extended target. Under the Gaussian assumption of $\overline{Q}_k$, the propagated distribution of the state in Eq. 17 remains Gaussian.

The temporal kernel is already trained by minimizing the log marginal likelihood of measurements using iteration methods like BFGS [19]. We used the Matern covariance function for the temporal part. To take a position and velocity as the dynamic of the targets, the one-time differentiable Matern $\nu = 3/2$ has been considered (Fig. 2.):

$$\kappa_t(\tau)_{\frac{3}{2}} = \sigma^2\left(1 + \frac{\sqrt{3}\tau}{l}\right)exp\left(-\frac{\sqrt{3}\tau}{l}\right), \tag{35}$$

where $\sigma, l$ are smoothness and length hyperparameters and $\tau = \Delta t$. The spectral density is:

$$S(\omega) = 1/(\lambda^2 + \omega^2)^2 \tag{36}$$

where $\lambda = \frac{\sqrt{3}\tau}{l}$. Thus, the spectral density can be factored as $S(\omega) = (\lambda + i\omega)^{-2}(\lambda - i\omega)^{-2}$. The transfer function of the corresponding stable part is $G(i\omega) = (\lambda + i\omega)^{-2}$. The continuous system matrix and the noise effect vector of the corresponding state-space model are derived as follows [10]:

$$\mathcal{A}_t = \begin{bmatrix} 0 & 1 \\ -\lambda^2 & -2\lambda \end{bmatrix}, \quad L_t = \begin{bmatrix} 0 \\ 1 \end{bmatrix}, \tag{37}$$
$$P_\infty = P_0 = \begin{bmatrix} \sigma^2 & 0 \\ 0 & \lambda^2\sigma^2 \end{bmatrix},$$

the spectral density of the Gaussian white noise process $w(t)$ is $Q_c = 4\lambda^3\sigma^2$ and the measurement model matrix is $H_t = [1\ 0]$. The state of each target can be defined as:

$$f_i(u^f) = [[f(u_1)_k, \dot{f}(u_1)_k], \ldots, [f(u_N)_k, \dot{f}(u_N)_k]]. \tag{38}$$

In a discrete case, $H_k = H_{\mathcal{T}}$, $L_k = L_{\mathcal{T}}$, $F_k = e^{\mathcal{A}_{\mathcal{T}}T_s}$ and stacked forms are:

$$\overline{x}_k = [f_1(u^f), f_2(u^f), \ldots, f_D(u^f)]^\top, \tag{39}$$
$$\overline{F}_k = I_D \otimes I_N \otimes F_k,$$
$$\overline{L}_k = I_D \otimes I_N \otimes L_{\mathcal{T}},$$
$$\overline{H}_k = I_D \otimes I_N \otimes H_{\mathcal{T}}.$$

Where $\otimes$ is the Kronecker product for stacked states, the subscript $\mathcal{T}$ denotes the temporal continuous model matrices in Eq. 39. The spectral density and the initialized state covariance have a spatial structure.

$$\overline{Q}_k = \overline{K_u} \otimes (P_\infty - F_kP_\infty F_k^\top), \quad \overline{P}_0 = \overline{K_u} \otimes P_{0,\mathcal{T}}, \tag{40}$$

In the case that the outputs are linearly dependent, we can say the $d$-th output of any input $u$ is expressed as a linear combination of $Q$ independent latent processes $U_q(u)$. For each output, we can write [17]:

$$f_d(u) = \sum_{q=1}^{Q} a_{d,q}U_q(u), \tag{41}$$

where each $U_q(u)$ represents an underlying latent process with different and independent GP priors. Here each latent function arises from a unique and independent latent GP. From Eq. 41, a computing covariance function between coupled outputs $\overline{f}(u) = [f_1(u), \ldots, f_D(u)]^\top$ can be expressed as:

$$cov(\overline{f}(u), \overline{f}(u')) = \sum_{q=1}^{Q} B_q\kappa_q(u, u'), \tag{42}$$

where $B_q = a_qa_q^\top$ is a positive definite (rank=1) matrix called a coregionalization matrix for $D$ outputs, and $\kappa_q(u, u')$ is the covariance function of the underlying process. Considering all the index points $u$ and using the Kronecker product, we can write the covariance matrix with the same prior kernel $Q = 1$ as:

$$\overline{K}(\overline{f}, \overline{f}') = B \otimes K_u(u, u'), \tag{43}$$

The Kronecker product of $B \otimes K$ can be written as Eq. 44. When $B$ is not an identity, the covariance of the joint process over $\overline{z}$ is not a diagonal matrix because of the dependency of the outputs. In fact, the observations of one target can affect the predictions for another target.

$$\overline{K}(\overline{f}, \overline{f}') = \begin{bmatrix} B_{11}K_u(u_1, u_1') & \ldots & B_{1D}K_u(u_D, u_D') \\ \ldots & \ddots & \ldots \\ B_{D1}K_u(u_1, u_1') & \ldots & B_{DD}K_u(u_D, u_D') \end{bmatrix}. \tag{44}$$

Using a stacked version of the last states, covariance and transition matrices in Eq. 39, we can predict the next state and covariance as follows:

$$\bar{\mathbf{x}}_{k+1|k} = \bar{\mathbf{F}}_k \bar{\mathbf{x}}_{k|k}, \tag{45}$$

$$\bar{\mathbf{P}}_{k+1|k} = \bar{\mathbf{F}}_k \bar{\mathbf{P}}_k \bar{\mathbf{F}}_k^\top + \bar{\mathbf{Q}}_k, \tag{46}$$

$$\bar{\mathbf{v}}_k = \mathbf{z}_k - \bar{\mathbf{H}}_k \bar{\mathbf{x}}_{k+1|k}, \tag{47}$$

$$\mathbf{S}_k = \bar{\mathbf{H}}_k \bar{\mathbf{P}}_{k+1|k} \bar{\mathbf{H}}_k^\top + \mathbf{R}_k, \tag{48}$$

$$K_k = \bar{\mathbf{P}}_{k+1|k} \bar{\mathbf{H}}_k^\top \mathbf{S}_k^{-1}, \tag{49}$$

$$\bar{\mathbf{x}}_{k+1|k+1} = \bar{\mathbf{x}}_{k+1|k} + K_k \mathbf{v}_k, \tag{50}$$

$$\bar{\mathbf{P}}_{k+1|k+1} = \bar{\mathbf{P}}_{k+1|k} - K_k \mathbf{S}_k K_k^\top. \tag{51}$$

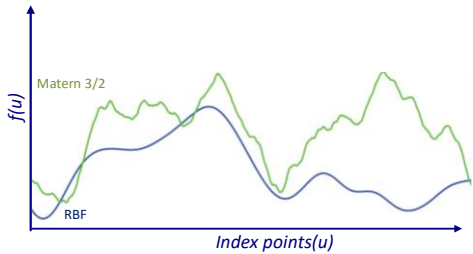

**FIGURE 3.** Matern 3/2 and RBF kernel function samples.

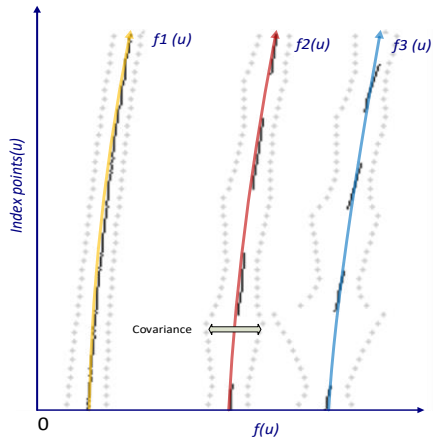

**FIGURE 4.** The extended target model for three lane lines. The measurements are depicted as black points, while the estimated curves are represented by colored points. The gray points illustrate the covariance range for each index.

## III. EXPERIMENTS AND EVALUATIONS

### A. Multi-Lane Tracking Application

This section describes the application of the proposed algorithm to track lane lines. The algorithm considers lane lines as extended targets that are dependent and dynamic smooth functions. A limited length vector value function and corresponding velocity vector are used to model the dynamic of each lane line. Due to the presence of challenging scenarios on the road, the environment is assumed to be cluttered. The measurements used in this application are video frames captured from a dashboard camera obtained from the CULane and TuSimple datasets [22]. The TuSimple dataset has about 7,000 one-second-long video clips of 20 frames each. They prepared the ground-truth result for the last frame (Frame 20). The TuSimple dataset covers most of the challenging situations on highways: curves, shadows, splitting, and merging. The relation between the measurements and the state of extended targets is depicted in

Fig. 3 using a limited number of index points $\mathbf{u_d}$, latent function or state $f_d(\mathbf{u})$ (colored points), their covariance (gray points), and the captured measurements (black points). For simplicity, the origin of the measurements is assumed to be the left bottom of the screen, similar to Fig. 3.

We utilized a subset of frames that contained annotated lane lines, depicted in Fig. 4, to train our model. We employed distinct time and space kernels and incorporated the target linear dependency via the coregonalization matrix B. Training was carried out using the Gpy Gaussian Process Python library and the BFGS optimization algorithm. The outcome of the training process is a set of hyperparameters that can be applied to calculate the state space transition matrices. In the spatial domain, we trained and optimized the RBF kernel, similar to Eq. 11, and in the temporal domain, the Matern kernel, equivalent to Eq. 35, using the training data $\{\mathbf{u}_d^*, \mathbf{z}_d^*\}_{d=1}^D$.

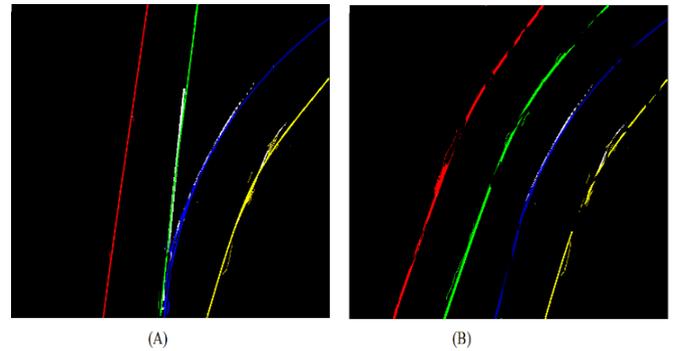

**FIGURE 5.** Labeled data used for training (A) Training data with shared measurements (B) Training data for parallel lane lines

---

**Algorithm 2. Training JPDACF**

**Input:** $\{\mathbf{u}_d^*, \mathbf{z}_d^*\}_{d=1}^D$, $\max_{\text{iters}}$, $K_{Type}$, $\sigma_0^2$, $l_0$, $\mathcal{T}$

**Output:** F, L, Qc, H, P_inf, P0, $B$

1: $B_0 = \text{Coregionalize}(D)$     #Initiation of $B$
2: $\kappa = B_0 \otimes K_{Type} (\sigma_0^2, l_0)$     #Initiate the kernel
3: model $\leftarrow$ Gpy.GPRegression ($\{\mathbf{u}_d^*, \mathbf{z}_d^*\}_{d=1}^D$, $\kappa$)
4: $B, l, \sigma^2 \leftarrow$ model.optimize('BFGS', max_iters)
5: **if** $K_{Type} = $ Matérn **then**
6:     $\lambda \leftarrow \sqrt{3}\tau/l$
7:     $Q_c \leftarrow 4\lambda^3 \sigma^2$
8:     $\mathcal{A}_t, L_t, P_\infty \leftarrow$ Eq. 37
9: **Endif** $K_{Type} = $ RBF **then**
10:     $\zeta \leftarrow \frac{1}{2l^2}$ , $N = 2$
11:     $Q_c \leftarrow \sigma^2 N! (4\zeta)^N \sqrt{\pi/\zeta}$
12:     $\mathcal{A}_t, L_t \leftarrow$ Eq. 14
13:     $P_\infty \leftarrow$ Eq. 15
14: **endif**
15: $H_k = H_\mathcal{T}$, $L_k = L_\mathcal{T}$, $F_k = e^{\mathcal{A}_\mathcal{T} T_s}$, $P_0 = P_\infty$
16: $Q_k = P_\infty - F_k P_\infty F^\top$
17: **return** $F_k, L_k, Q_k$, H, P0, $B$

---

In Fig. 5, the results of the dependency matrix from the training data are displayed. In each captured frame, the process of preparing measurements involves normalizing the picture and identifying features associated with each target.

For lane-line detection applications similar to those in [23] and [24], edge and color features are utilized. Additional preprocessing algorithms, such as determining the region of interest (ROI) and removing other objects from the scene, are essential. Additionally, to eliminate any camera perspective effects, regular Inverse Perspective Mapping (IPM) is applied, as demonstrated in [25] and [26]. This facilitates easier modeling of the dependency of features in lane lines.

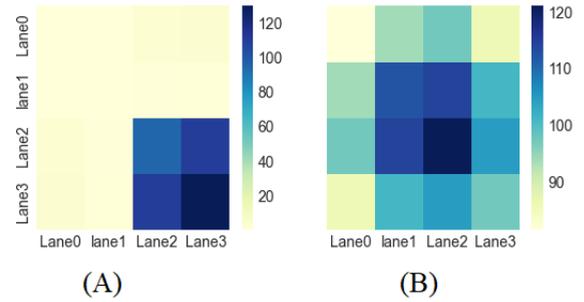

(A)  (B)

**FIGURE 6.** The Heat map of the dependency matrix for training data is in Fig.4. In subfigure (A), the dependencies between lanes 0 and 1, as well as 2 and 3, are illustrated such that the closer intensity in the heat map shows more dependency. In Subfigure B, all lanes are dependent and have almost similar numbers.

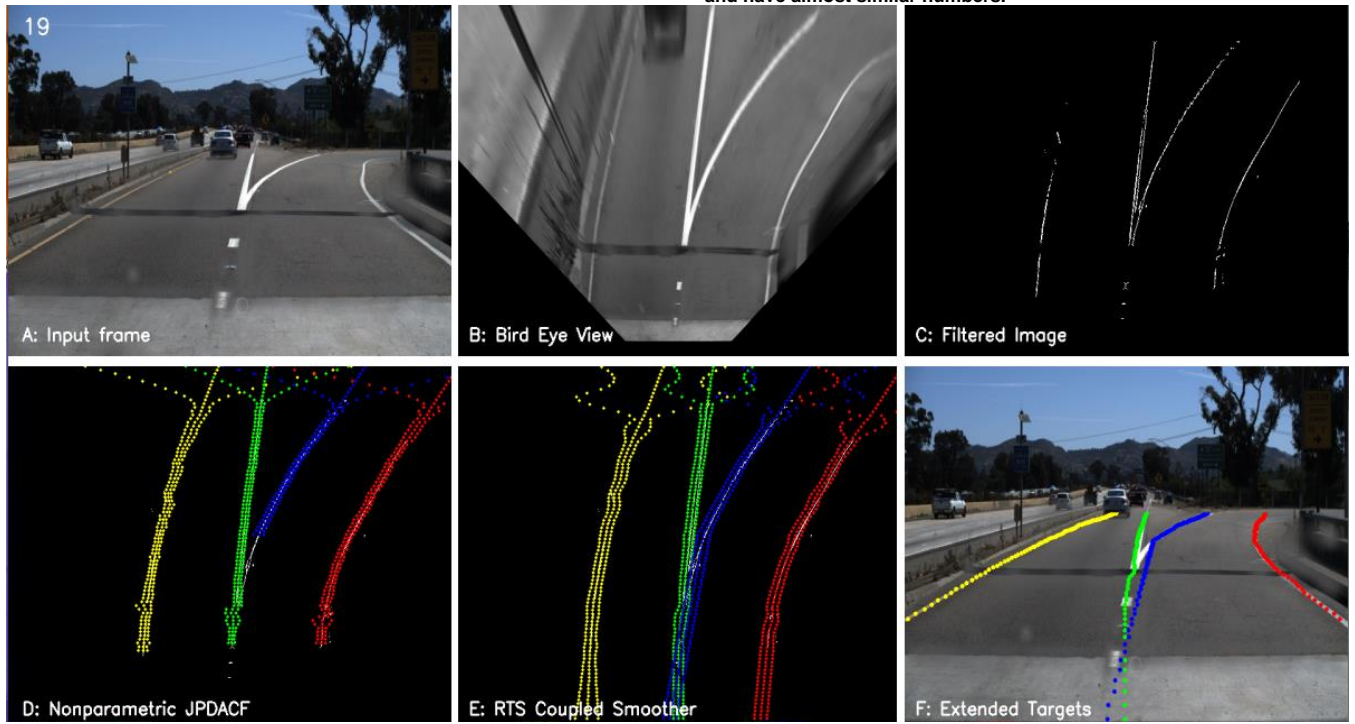

A: Input frame  B: Bird Eye View  C: Filtered Image

D: Nonparametric JPDACF  E: RTS Coupled Smoother  F: Extended Targets

**FIGURE 7.** Comparison of the proposed approach with SCNN fully supervised method.

The processing steps are illustrated in Fig.6, including:

    A. Read frames from a video.
    B. Create a Bird's Eye View.
    C. Apply feature filtering (edge and color).
    D. Use JPDACF to identify measurements.
    E. Use the RTS smoother to calculate the full posterior.
    F. Track the extended targets.

We compared our algorithm with a fully supervised lane detection method recently published in [14]. In this paper, the Spatial Convolutional Neural Network (SCNN) was used to detect lane lines; the information on the spatial correlation between rows and columns for detection was extracted. This method achieved good results when detecting lane markings in the presence of clutters, especially when the lane lines are straight. The approach uses probability maps (probmaps) of lane markings resulting from the SCNN and, after verification, fills the maps with a cubic spline, which makes the final prediction. The SCNN works well when the lanes are straight with enough feature cues, but in situations with a lack of distinctive features, it produces miss clustering and incorrect detection as in Fig. 7. This occurs most frequently at curvature-type lane-lines, splitting, and merging when the symmetric structure is missed. We tested our algorithm against the SCNN algorithm in merging and splitting videos in the TuSimple dataset. We tested on five different videos, amounting to a total of 100 frames. Our method yielded more that 20% improvement in accuracy, false positive, and false negative rates, respectively, in merging and splitting situations over the SCNN method. We used the same accuracy formula as the TuSimple benchmark: space model representation.

$$accuracy = \frac{\sum_{clip} C_{clip}}{\sum_{clip} S_{clip}},\qquad(52)$$

where $C_{clip}$ is the number of correct points in the last frame of the clip, and $S_{clip}$ is the number of requested points in the last frame of the clip. If the difference between the distance of the ground truth and the prediction is less than a threshold, the predicted point is a correct one.

Based on Eq. 52 and Table II, we also computed the rate of false positives and false negatives for the test results. False positive means the lane is predicted but not matched with any lanes in the ground truth. False negative means that the lane is in the ground truth but not matched with any lanes in the prediction.

**TABLE II.** Result of SCNN algorithm in Merging and Splitting situations.

|  | SCNN Method in [14] | | | | Proposed Method | | | |
|---|---|---|---|---|---|---|---|---|
|  | FPR | FNR | Accuracy | Frame/s | FPR | FNR | Accuracy | Frame/s |
| Merging and Splitting | 0.350 | 0.266 | 0.877 | 0.710 | 0.183 | 0.183 | 0.895 | 5.0 |

$$FP = \frac{F_{pred}}{N_{pred}}, FN = \frac{M_{pred}}{N_{gt}}, \qquad (53)$$

where $F_{pred}$ is the number of wrongly predicted lanes, $N_{pred}$ is the number of all predicted lanes. $M_{pred}$ is the number of missed ground-truth lanes in the predictions, and $N_{gt}$ is the number of all ground-truth lanes.

We also compared our running time with the SCNN algorithm on a normal CPU. Our algorithm's run-time in a computer with an Intel i7 CPU, clocked at 2.9 GHz with 16 GB RAM for those scenarios, was about five times faster than SCNN, as presented in [14]. The results are shown in Table II.

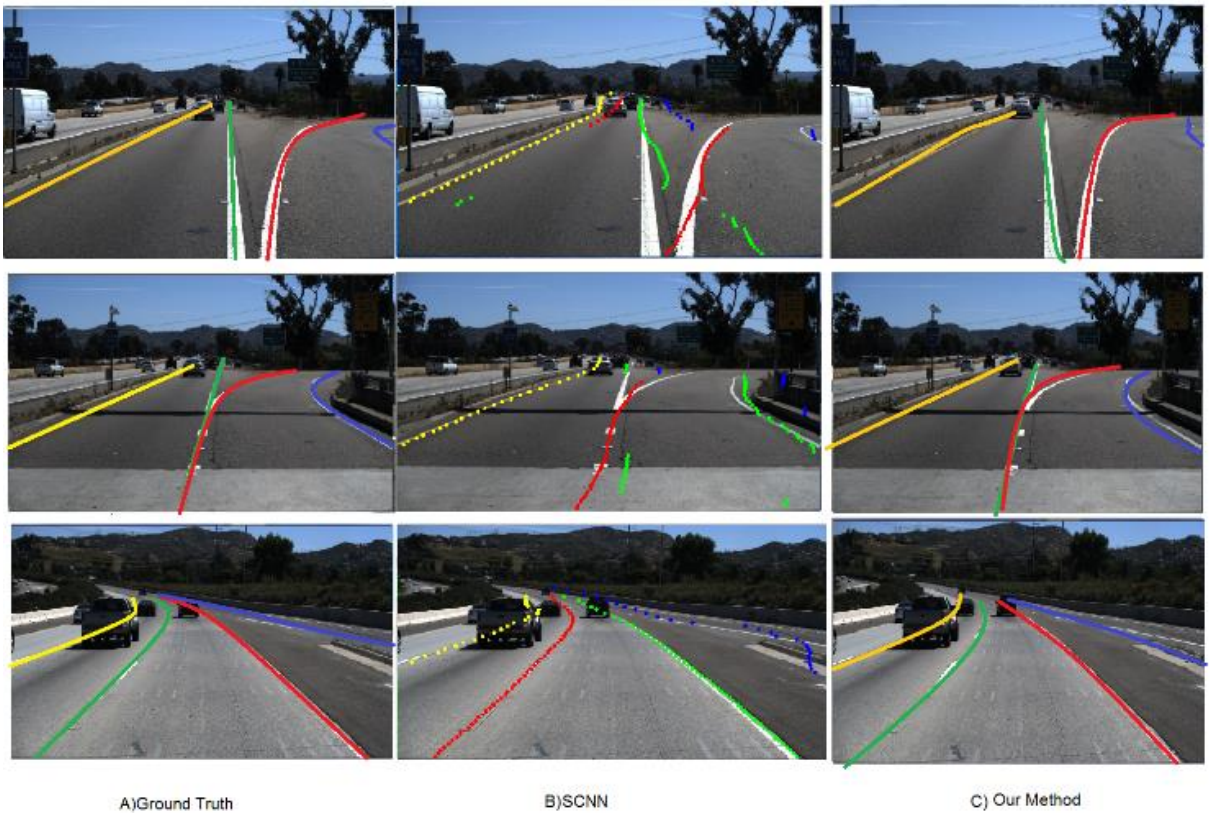

A)Ground Truth     B)SCNN     C) Our Method

**FIGURE 8.** Comparison of the proposed approach with SCNN fully supervised method.

## IV. CONCLUSION

This research proposes an ST-JPDACF algorithm for tracking multiple extended targets. The algorithm is divided into two tracker algorithms, one in the space domain for predicting measurements and corresponding likelihoods using nonparametric JPDACF, and the other in the time domain for Bayesian multivariate tracking. The use of JPDACF enables the algorithm to handle coupling, clutters, and assignment uncertainties. By employing various trainable kernel functions, we were able to manage the dependency of measurements in space (within a frame) and time (between frames), which can be learned from a limited amount of training data. This extension can be utilized to track the shape and dynamics of nonparametric dependent extended targets in the presence of clutter, even when targets share measurements.

For future work, the proposed algorithm could be used to model other measurement models, such as 3D non-star convex objects. Additionally, applying GP trajectory optimization using a Factor graph and its application in extended target tracking could be explored.

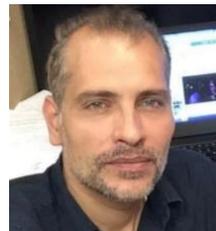

**Behzad Akbari** , a member of IEEE, obtained his bachelor's and master's degrees in computer software and computer architecture from Tehran and Knowledge And Research University, Iran, in 1996 and 1999 respectively. He later completed his second master's and PhD degrees in Computer Science and Electrical and Computer Engineering (ECE) from McMaster University, Canada, in 2014 and 2021 respectively. Currently, he holds a postdoctoral fellowship at Dalhousie University, Halifax, Nova Scotia, Canada, in the Department of Mechanical Engineering. He has editorial experience with IEEE Access, Transactions on Robotics, and Transactions on Systems, Man and Cybernetics (SMC). His research focuses on state estimation algorithms, collaborative multi-agent systems, multi-target tracking, multi-output Gaussian process, iterative localization and mapping, factor graph optimization, and reinforcement learning.

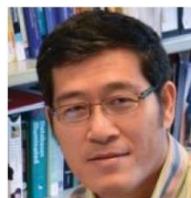

**Haibin Zhu (M'02–SM'04)** received his B.S. degree in computer engineering from the Institute of Engineering and Technology, Zhengzhou, China (1983), and M.S. (1988) and Ph.D. (1997) degrees in computer science from the National University of Defense Technology (NUDT), Changsha, China.He is a Full Professor and the Coordinator of the *Computer Science program,* Founding Director of the *Collaborative Systems Laboratory, Nipissing University, Canada*. He was a visiting professor and a special lecturer in the College of Computing Sciences, New Jersey Institute of Technology, USA (1999-2002) and a lecturer, an associate professor and a full professor at NUDT (1988-2000). He has accomplished (published or in press) over 200 research works including 30+ IEEE Transactions articles, six books, five book chapters, three journal issues, and three conference proceedings.

He is a fellow of ICIC (Institute of Cognitive Informatics and Cognitive Computing), a senior member of ACM, a senior member of IEEE, a full member of Sigma Xi, and a life member of CAST-USA (Chinese Association of Science and Technology, USA). He is serving as a member-at-large of the Board of Governors (2022-), Vice President, Systems Science and Engineering (SSE) (2023-) and a co-chair (2006-) of the technical committee of Distributed Intelligent Systems of IEEE Systems, Man and Cybernetics (SMC) Society (SMCS), Editor-in-Chief of IEEE SMC Magazine (2022), Associate Editor (AE) of IEEE Transactions on SMC: Systems (2019-), IEEE Transactions on Computational Social Systems(2019-), Frontiers of Computer Science (2021-), and IEEE Canada Review (2019-). He was AE of IEEE SMC Magazine (2018-2021), Associate Vice President (AVP), SSE (2021), IEEE SMCS, a Conference (Co-)Chair and Program (Co-)Chair for many international conferences, and a PC member for 130+ academic conferences. He is the founding researcher of Role-Based Collaboration and the creator of the E-CARGO model. His research monograph E-CARGO and Role-Based Collaboration can be found https://www.amazon.com/CARGO-Role-Based-Collaboration-Modeling-Problems/dp/1119693063. The accompanying codes can be downloaded from GitHub: https://github.com/haibinnipissing/E-CARGO-Codes. He has offered 20 keynote and plenary speeches for international conferences and 80 invited talks internationally. His research has been being sponsored by NSERC, IBM, DNDC, DRDC, and OPIC. He is the recipient of the best paper award in international collaboration from the 25th Int'l conf. on Computer-Supported Cooperative Work in Design, Hangzhou, China, 2022, the meritorious service award from IEEE SMC Society (2018), the chancellor's award for excellence in research (2011) and two research achievement awards from Nipissing University (2006, 2012), the IBM Eclipse Innovation Grant Awards (2004, 2005), the Best Paper Award from the 11th ISPE Int'l Conf. on Concurrent Engineering (ISPE/CE2004), the Educator's Fellowship of OOPSLA'03, a 2nd class National Award for Education Achievement (1997), and three 1st Class Ministerial Research Achievement Awards from China (1997, 1994, and 1991). His research interests include Collaboration Systems, Human-Machine Systems, Computational Social Systems, Collective Intelligence, Multi-Agent Systems, Software Engineering, and Distributed Intelligent Systems.

s

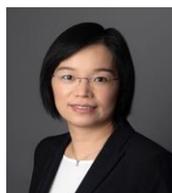

**Ya-Jun Pan (S'00-M'03-SM'11)** is a Professor in the Dept. of Mechanical Engineering at Dalhousie University, Canada. She received the B.E. degree in Mechanical Engineering from Yanshan University, the M.E. degree in Mechanical Engineering from Zhejiang University, and the Ph.D. degree in Electrical and Computer Engineering from the National University of Singapore. She held post-doctoral positions of CNRS in the Laboratoire d'Automatique de Grenoble in France and the Dept. of Electrical and Computer Engineering at the University of Alberta in Canada respectively. Her research interests are robust nonlinear control, cyber physical systems, intelligent transportation systems, haptics, and collaborative multiple robotic systems. She has served as Senior Editor and Technical Editor for IEEE/ASME Trans. on Mechatronics, Associate Editor for IEEE Trans. on Cybernetics, IEEE Transactions on Industrial Informatics, IEEE Industrial Electronics Magazine, and IEEE Trans. on Industrial Electronics. She is a Fellow of Canadian Academy of Engineering (CAE), Engineering Institute of Canada (EIC), ASME, CSME, and a registered Professional Engineer in Nova Scotia, Canada.

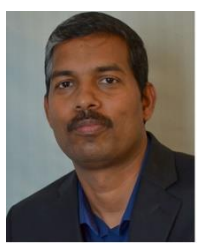

Ratnasingham Tharmarasa, a Senior Member of IEEE, completed his bachelor's degree in electronic and telecommunication engineering at the University of Moratuwa, Sri Lanka, in 2001. He further pursued his master's and doctoral degrees in electrical engineering at McMaster University, Canada, in 2003 and 2007 respectively. Currently, he serves as an Assistant Professor in the Department of Electrical and Computer Engineering at McMaster University. From 2001 to 2002, he worked as an Instructor of Electronic and Telecommunication Engineering at the University of Moratuwa. From 2002 to 2007, he was a Graduate Student/Research Assistant at the ECE Department of McMaster University, and later worked as a Researcher with DRS Technologies Canada Limited in 2008. From 2008 to 2019, he held the position of Research Associate at the ECE Department. He has authored or coauthored over 50 peer-reviewed journal articles, 65 conference papers, and three book chapters in his areas of expertise. Additionally, he has been involved in the development of real-world target tracking, sensor fusion, and resource management systems. His research interests include target tracking, information fusion, sensor resource management, and performance prediction, with applications in surveillance systems, autonomous vehicles, and intelligent transportation..